\documentclass[runningheads]{llncs}

\usepackage[T1]{fontenc}
\usepackage[utf8]{inputenc}
\usepackage{graphicx}
\usepackage{amsmath,amssymb}
\usepackage{booktabs}
\usepackage{multirow}
\usepackage{makecell}
\usepackage{float}
\usepackage{colortbl}
\usepackage{xcolor}
\usepackage{subfigure}
\usepackage{url}
\usepackage[hypertexnames=false]{hyperref}

\definecolor{mygray}{gray}{.9}

\begin{document}

\title{Understanding and Tackling Label Errors in Individual-Level Natural Language Understanding}
\titlerunning{Label Errors in Individual-Level NLU}

\author{Yunpeng Xiao\inst{1} \and Kyrie Zhou \inst{2} Youpeng Zhao\inst{3} \and Kai Shu\inst{1}}
\authorrunning{Y. Xiao et al.}
\institute{Emory University, Atlanta, GA, USA\\
\email{\{yunpeng.xiao,kai.shu\}@emory.edu}
\and
University of Texas at San Antonio, San Antonio, TX, USA\\
\email{kyrie.zhou@utsa.edu}
\and
University of Central Florida, Orlando, FL, USA\\
\email{youpeng.zhao@ucf.edu}}

\maketitle

\begin{abstract}
Natural language understanding (NLU) is a task that enables machines to understand human language. 
Some tasks are closely related to individual subjective perspectives, thus termed individual-level NLU. 
Previously, these tasks are often simplified to text-level NLU tasks, ignoring individual factors such as demographic information or worldview over a period of time. This can lead to a large number of label errors and also make inference difficult and unexplainable.
To address the above limitations, we propose a new NLU annotation guideline based on individual-level factors. 
We find that the error rate in the dataset samples exceeded 20\%, with the highest reaching 48.7\%.
We further use large language models to conduct experiments on the re-annotation datasets and find that the large language models perform well on these datasets after re-annotation. We also verify the effectiveness of individual factors through ablation studies. Our re-annotation dataset can be found at \href{https://anonymous.4open.science/r/Individual-NLU-A0DE}{https://github.com/24yearsoldstudent/Individual-NLU}.

\end{abstract}

\section{Introduction}

\begin{figure}[t]
	\centering
	\subfigure[The original dataset contains text, target and label. LLM judge on the text is different from the label.]{
		\begin{minipage}{7.5cm}
                        \includegraphics[width=\textwidth]{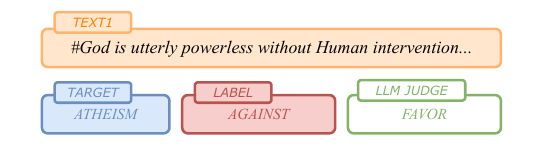} \\
		\end{minipage}
	}

	\subfigure[Expanded dataset, add other posts of the same individual (user) for the same target, and re-annotate the user's stance. The LLM Judge is consistent with the re-annotation label.]{
		\begin{minipage}{7.5cm}
			\includegraphics[width=\textwidth]{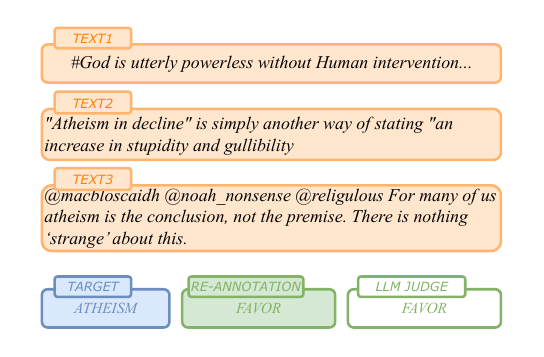} \\
			
		\end{minipage}
	}
    \caption{A typical example of potential label error in stance detection.}
    \label{label_errors}
    
\end{figure}

Natural language understanding (NLU) is the task that uses semantic and syntactic analysis to enable computers to understand human-language inputs. Common natural language understanding tasks include fake news detection \cite{shu2017fake}, sentiment analysis \cite{wankhade2022survey}, stance detection \cite{aldayel2021stance}, toxicity/hate speech detection \cite{pavlopoulos2020toxicity}, and sarcasm detection \cite{joshi2017automatic}. Existing NLU datasets are predominantly text-based, relying solely on short text information without accounting for social factors. While text-level NLU simplifies many tasks, its limitations begin to be recognized, such as poor inference performance \cite{hovy2021importance,bhattacharya2025rethinking}.
So, some researchers have propose frameworks integrating social factors into NLU \cite{hovy2021importance}. Additionally, various studies have incorporated different social factors such as user information and background knowledge into specific tasks to improve NLU accuracy \cite{yang2017overcoming,aldayel2019your}.

However, current research has not explored in depth which NLU tasks will have huge deficiencies when using only textual information (without any other factors). In this paper, we define a type of NLU tasks as individual-level NLU tasks, where the labels reflect the identity or perspective of the individual (typically the web user who posts the text) rather than the content of the text itself. 
We argue that inference only with short texts is flawed in such tasks. Tasks that fall under individual-level NLU include sentiment analysis, sarcasm detection, and stance detection etc. 
A key characteristic of these tasks is that their labels are inherently tied to the publishers rather than the readers, or at least the publisher's identity and perspective play an important role in the label.
An NLU task that does not incorporate an individual's perspective is not considered individual-level NLU. Such tasks are usually annotated based on social consensus or objective facts.
For example, in tasks such as natural language inference, the labels usually represent a broadly accepted interpretation rather than an individual user's perspective \cite{bowman2015large}. 
In fake news detection \cite{shu2017fake} and authorship detection \cite{huang2024can}, the label remains unchanged regardless of whether one or multiple individuals share or endorse it. 

This deficiency is reflected in the creation of the datasets. Current research often implicitly assumes that the labels in original datasets are accurate. 
However, individual-level NLU datasets are often created using text-level guidelines, and annotators' interpretations may differ from those of the original publishers. 
Such misalignment can lead to a significant number of labeling errors. 
Prior works have attempted to mitigate this issue by leveraging the individual factors in the datasets. 
For instance, datasets like Amazon reviews \cite{zhang2015character} and IMDB \cite{maas2011learning} assign labels directly based on user scores, reducing the likelihood of annotation inconsistencies. However, many individual-level NLU datasets, such as the Twitter stance detection dataset \cite{mohammad2016semeval} and the Twitter sentiment analysis dataset \cite{rosenthal2019semeval}, depend mostly on manual annotation, since social media posts do not come with explicit "scores" and must be annotated manually or inferred through hashtags instead. A recent study demonstrates that large language models (LLMs) perform well when human annotators do but fail in cases where human annotators struggle to reach consensus \cite{li2024advancing}. This suggests that inconsistencies among annotators stem from the inherent ambiguity of the text rather than from annotator negligence. 

From a sociolinguistic perspective, the attitude of an individual should be tied to the original publisher’s intent at the time of posting \cite{kockelman2004stance}, rather than being subject to the variability of annotator interpretations. Annotation inconsistencies often arise due to insufficient information and poor data quality. This phenomenon is referred to as systematic label errors \cite{cabrera2014systematic}. Prior work has reported non-trivial label error even under the original text-level stance detection; the error rate for the atheism category in the SemEval-2016 dataset was as high as 22.7\% \cite{garg2024stanceformer}. Beyond such mislabels, we argue that a single post is often informationally insufficient for user-level inference, motivating the use of multi-post evidence. We later validate this necessity via a simple ablation on the amount of user context.

To address this research gap, we propose guidelines for three NLU subtasks: stance detection, topic-based sentiment analysis, and hate speech detection. These guidelines aim to identify and mitigate systematic labeling errors that may exist in text-level NLU datasets. Specifically, we incorporate additional posts from the same user within a similar timeframe to assess the accuracy of dataset labels. We mitigate these errors by incorporating additional posts from the same user within a bounded time window, using cross-post consistency as evidence for the individual-level label. Our analysis reveals a substantial number of labeling errors. To further evaluate these errors, we employ three mainstream large language models (LLMs) to evaluate the datasets. Our findings indicate that LLMs achieve exceptionally high accuracy on the re-annotated datasets using only simple prompts, demonstrating the necessity of introducing individual-level NLU and individual factors.
 
We summarize our contributions as follows:
\begin{itemize}
    \item We propose a novel guideline to reduce labeling errors in individual-level NLU. 
    \item We show that widely used individual-level benchmarks can substantially underestimate model performance due to systematic label errors, and that evaluation on our corrected labels yields markedly higher and more consistent scores across models.
    \item We show that using multi-post user context is crucial for user-level inference, supported by controlled ablations on the amount of available context.
\end{itemize}

\section{Related Work}
In this section, we will introduce existing the definition for label errors. Then we will introduce pre-trained and large language models, and explain their potential in detecting label errors in the individual-level NLU task.

\subsection{Label Errors}
The inconsistency between the labels and groundtruths in the training dataset is often called "noisy labels" \cite{song2022learning}. If the labels are inconsistent with the groundtruths in the test dataset, it is called label errors. Label errors are common in test datasets and may affect the evaluation of the model. There is an average of 3.3\% label error in ten commonly used datasets \cite{northcutt2021pervasive}. 

Large number of label errors (over 20\%) are probably not due to the negligence of the annotators but the defects in the annotation guidelines themselves. Label errors at such scale also imply pervasive noisy labels in the training data, biasing both training and test distributions away from the true data-generating process and thereby undermining classical label-error detection methods that assume a relatively clean test set or stable noise patterns. For example, the 23.7\% label error rate in the TADRED dataset is because of inappropriate guidelines \cite{stoica2021re}. Annotation guidelines serve as the instruction manual for annotators, drafted by product owners. The process can be simply summarized as follows: (1) Annotators are recruited and given data samples and the description of guidelines; (2) Annotators provide the labels based on their knowledge and experience, by strictly complying with the guidelines \cite{klie2024analyzing}.

\subsection{Pre-trained Language Models}
Before the emergence of large language models, studies have shown that pre-trained language models are better than support vector machines or other deep learning models \cite{ghosh2019stance}. Many works demonstrate that using external knowledge can effectively enhance the performance of individual-level NLU tasks such as stance detection tasks \cite{he2022infusing,hanawa2019stance,li2021improving}. Since large language models were pre-trained with a large corpus, many researchers began to explore their performance in individual-level tasks such as stance detection \cite{zhang2022would,cruickshank2023use,lan2024stance,li2024advancing,gatto2023chain}, sentiment analysis \cite{zhang2023sentiment,korkmaz2023analyzing}.  However, these works focus on how to guide LLMs to achieve better performance, and no work has considered the potential impact of label errors in individual-level NLU tasks. If the dataset is systematically and consistently mislabeled, the evaluation of LLMs can become both misleading and unreliable.

\section{Methodology}
In this section, we illustrate the process of mitigating label errors in individual-level NLU tasks. We begin by highlighting the unique characteristics of individual-level tasks. Next, we present our methods, using representative tasks such as stance detection and topic-based sentiment analysis.

\begin{figure*}
    \centering 
    \includegraphics[width=1\textwidth]{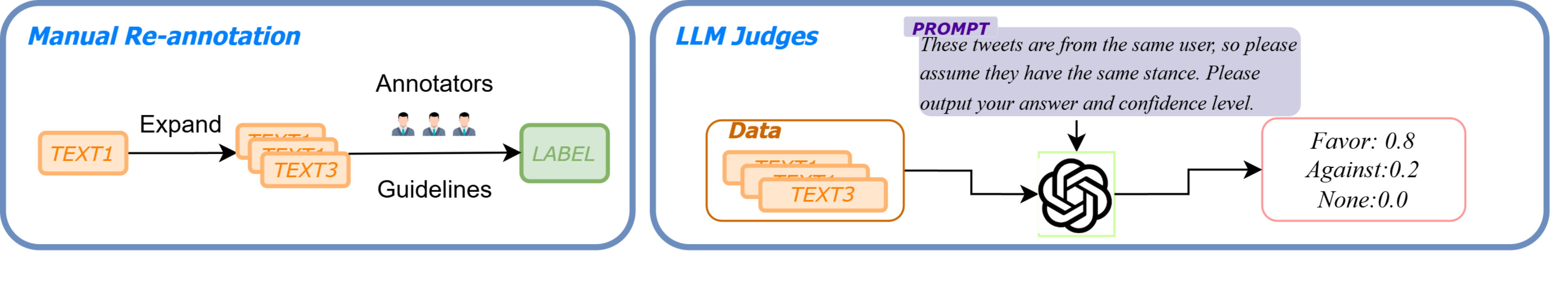} 
    \caption{The process of manual re-annotation and LLM judges. In the manual re-annotation, after finding other posts related to the topic/target by individuals (network users), three annotators follow the guidelines to annotate individual-level labels. In LLM judges, the input is divided into two parts: data and prompts.} 
    ~\label{reannotation_process} 
    \vspace{-5mm}
\end{figure*}

\vspace{-2mm}

\subsection{Tasks and Dataset Selection}
According to the definition of individual-level NLU, it's difficult for annotators to directly infer a publisher's perspectives, but they can approximate them using indirect contextual information about the user. Relying solely on a single piece of text often results in inaccurate annotations. This highlights the critical need for a more comprehensive understanding of an individual’s background in NLU tasks, including physiological attributes (e.g., gender, age) and social factors (e.g., interests, occupation, and community affiliations). However, collecting such sensitive information from social media presents significant challenges, particularly regarding privacy concerns. Therefore, it is essential to simplify the problem by focusing on specific individual-level NLU tasks while minimizing privacy risks.

Therefore, we focus on three representative tasks: topic-based sentiment analysis, stance detection, and hate speech detection. One advantage of these tasks and datasets is that they have clearly defined topics or targets, making it easier to collect relevant posts from the users. Other tasks, such as sarcasm detection lack a specific target and often require a deep understanding of an individual’s speaking style, interests, and other contextual factors, making data collection significantly more challenging. Additionally, previous studies have shown that users' stances and sentiment toward specific perspectives tend to remain stable over short periods \cite{borge2015content,aldayel2019your}.  For example, in topic-based sentiment analysis, if the topic is Arsenal, a dedicated Arsenal fan is expected to maintain a positive sentiment toward the team over time. For our study, we select three datasets: the SemEval-2016 Task 4 topic-based sentiment analysis dataset \cite{nakov2019semeval}, the SemEval-2016 Task 6 stance detection dataset \cite{mohammad2016semeval}, and the HateXplain dataset \cite{mathew2021hatexplain}. The basic statistics of three datasets are presented in Table \ref{demographic}. Stance detection has three classes: \textsc{Favor}, \textsc{Against}, and \textsc{None}. In topic-based sentiment analysis, the creators discard the $Neutral$ and only keep the \textsc{Positive} and \textsc{Negative} classes. HateXplain also has three classes, namely $Hatespeech$, \textsc{Offensive}, and \textsc{Normal}. 

\begin{table}[t]
\centering
\caption{Statistics of three datasets.} ~\label{demographic}
\begin{tabular}{llll}
\hline
Dataset & Topics/Targets & Tweets      \\ \hline
Semeval Stance & 5 & 1,129  \\
\hline
Semeval Sentiment & 60 & 4,346    \\
\hline
HateXplain & 12 & 20,148    \\
\hline
\end{tabular}
\vspace{-5mm}
\end{table}

\subsection{Data Expansion}
To expand the dataset, we collect user posts related to the specified topic or target. We make the following assumption: given a set of posts $X = {x_1, x_2, ..., x_n}$ authored by a user about a topic or target $t$ over a certain period, these posts should exhibit the same perspective. We further validate this assumption from a clustering perspective. Previous research has clustered posts based on textual features at the text level \cite{samih2021few}, where posts with similar textual characteristics are positioned closer together and are more likely to share the same class label. At the individual level, drawing from prior studies \cite{borge2015content,aldayel2019your}, we extend this idea by assuming that users and their posts should be each other's nearest neighbors. In other words, if a dataset contains only one post $x_1$ from a given user, and we add $k$ additional posts ${x_1,...x_k}$, forming a cluster of nearest neighbors that share the same label. Previous studies have shown that detecting label errors requires as few as two nearest neighbor samples (2-NN) \cite{zhu2022detecting}.  so we set $k=2$ for the dataset creation (we also conduct experiments to demonstrate the impact of $k$ in section 6.2). However, certain edge cases must be considered—such as when a user has only one post related to $t$, or when the original post itself isn't directly related to $t$. We provide specific guidelines for handling the cases in Section 3.3.

We start from the existing dataset, find the users corresponding to these posts, and then use the Twitter API to crawl other tweets from the same user within a certain period of time based on the corresponding keywords. This period is usually no more than two years. For example, for the stance dataset, we crawl the user’s tweets from January 2015 to December 2016. If more than three tweets are collected from a user, we filter the tweets: We first keep the tweets that explicitly contained the target (e.g., the target was Legalization of Abortion and the tweet explicitly contained abortion). If there are not enough tweets (less than three tweets), we manually collect the user's tweets in the following order: (1) tweets posted by the user related to the target or topic (regardless of time, the closer to the original tweet, the better); (2) tweets retweeted by the user related to the target or topic (regardless of time, the closer to the original tweet, the better); (3) tweets posted by the user closest to the original tweet. 

Because some tweets have been deleted or restricted, we can only obtain the users corresponding to a part of the tweets. This is also the case in previous studies \cite{aldayel2019your}. In the stance detection dataset, we selected four targets: Atheism (AT), Climate Change is a Real Concern (CC), Feminist Movement (FM), and Legalization of Abortion (LA) for data expansion. This is because worldview-anchored targets (e.g., atheism, feminism, abortion) than for candidate-centric targets, where attitudes can shift rapidly in response to campaign events or news \cite{sagiv_schwartz_2022_values}. Therefore, we focus our re-annotation on the former and exclude politician targets (Hillary Clinton / Donald Trump) from this study. In topic-based sentiment analysis, we select two tweets for each topic, giving priority to one with the label \textsc{Positive} and one with the label \textsc{Negative}. However, if all tweets of a certain class of a certain topic are inaccessible, we select two tweets of the same class. The data statistics are shown in Table \ref{tab:error_rate}. In HateXplain, each tweet has three annotators, and we select tweets with inconsistent annotators in the test set (e.g., two annotators annotated “Hatespeech” and one annotator annotated “Offensive”). There are 350 such tweets in total, but only 76 of them are still accessible. Our data expansion assumes that a user’s stance toward a target is approximately stable within a bounded time window. 

\subsection{Manual Re-annotation Guidelines}
After collecting user tweets, three annotators (all of them are authors) independently annotated them. Considering that we use LLMs for data evaluation and that the annotators may not understand some background knowledge, we allow the annotators to use search engines to assist in the annotation work but prohibit the use of LLMs. 

In the annotation, we first followed the guidelines for three datasets. According to the characteristics of expanded data, we propose a new guideline for stance detection and topic-based sentiment analysis: for individuals whose sentiment/stance is difficult to determine, we use the following rules to annotate: (1) If none of the three posts can determine the individual's stance/sentiment on the target, it is annotated as \textsc{None} ($Neutral$). (2) If one tweet clearly states that the stance is \textsc{Favor} (\textsc{Positive}) or \textsc{Against} ($Negetive$), and the remaining two tweets have unclear stances or are irrelevant to the target, the stance is still annotated as \textsc{Favor} (\textsc{Positive}) or \textsc{Against} ($Negetive$). (3) If more than half of the tweets are \textsc{Favor} (\textsc{Positive})/\textsc{Against} ($Negetive$), please identify the user's stance/sentiment as \textsc{Favor} (\textsc{Positive})/\textsc{Against} ($Negetive$). 

The situation for the HateXplain dataset is more complicated. Hate speech detection is a complex task that combines individual factors and social consensus. First, determining whether a speech is offensive is mainly based on the perspective of readers/social consensus, while determining whether a speech is hate speech requires considering both individual factors and social consensus. Second, unlike stance detection and sentiment analysis tasks, if a post is judged as hate speech, it does not mean that other posts of the user are also hate speech; however, the individual level obtained from other posts can play an auxiliary role in judging whether the current post is hate speech. In fact, there are differences in the definition of hate speech and even the annotation guidelines of different datasets \cite{jahan2023systematic}. In HateXplain, $Hatespeech$, \textsc{Offensive}, and \textsc{Normal} also rely mainly on the subjective judgment of the annotator, and there is no standard. Compared with the other two tasks, the fuzzy label boundaries of hate speech detection exacerbate the uncertainty of labels. We used the United Nations definition of hate speech \cite{strossen2021united} in re-annotation and developed the standards for distinguishing three classes of labels. These standards are also input into LLMs as system prompts. 

The Krippendorff's $\alpha$ values for stance detection, topic-based sentiment analysis, and hate speech detection datasets were 0.68, 0.71, and 0.52, respectively. Clearly, the ambiguity of the hate speech detection boundaries leads to greater inconsistencies. However, our goal is to re-annotate the labels of the original dataset; we cannot simply discard the samples when the three annotators are inconsistent. Therefore, if an inconsistency is found, the annotators will re-search the Twitter user's information and discuss it until they reach a consensus. 
\begin{figure*}[t]
    \centering 
    \includegraphics[width=1\textwidth]{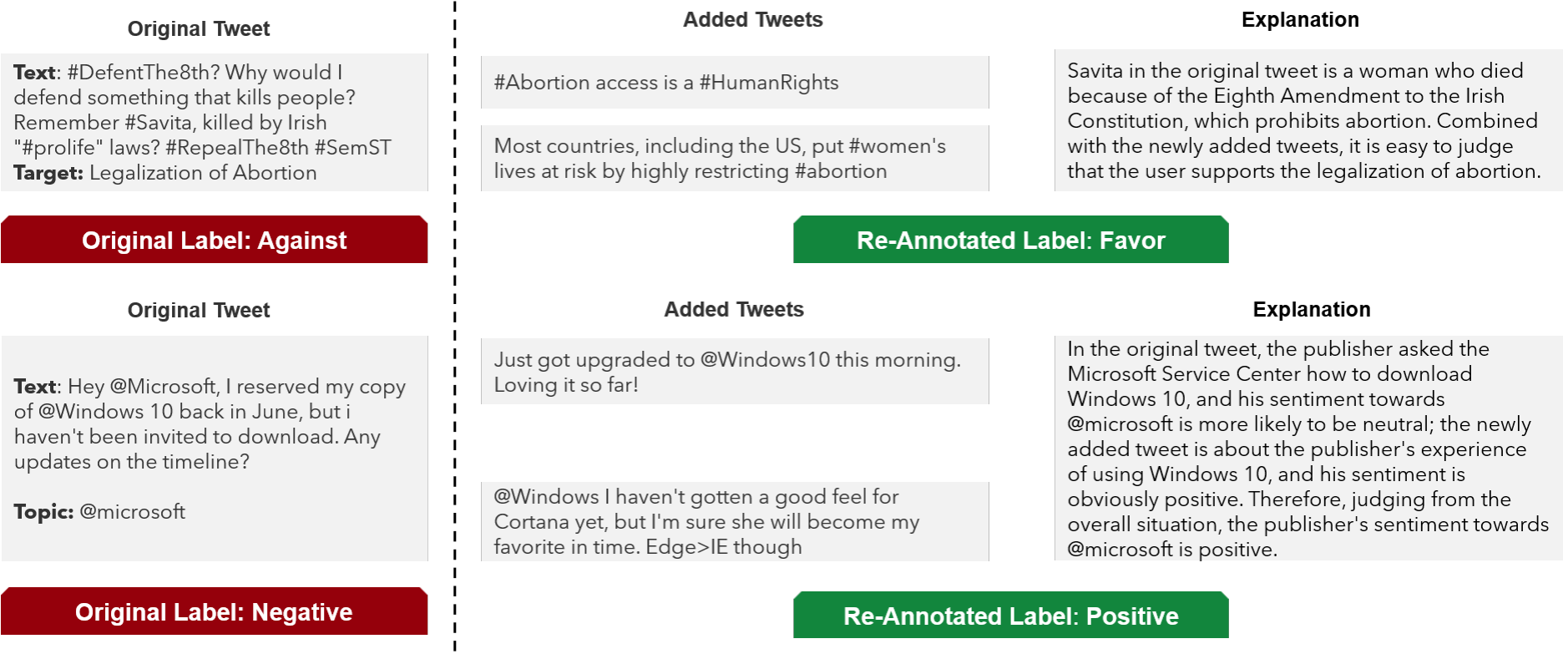} 
    \caption{Some examples of correcting label errors. Using multiple posts from the same publisher can more accurately determine the user's sentiment or stance, and can effectively explain why this label is given.} 
    ~\label{error_simples} 
\end{figure*}
\section{Experiments}
In this section, we introduce the large language models used to evaluate Individual-level NLU performance and then our evaluation metrics. 

\subsection{LLM Evaluations}

We use three representative large language models: GPT-4o \cite{achiam2023gpt}, Llama3-70B \cite{dubey2024llama}, and PHI-4 \cite{abdin2024phi} to evaluate the performance of the datasets after expansion and correction. 


To demonstrate that multiple posts are more effective than one post, we also conduct two ablation experiments. The first ablation experiment compares the performance when using the original tweet and two newly collected tweets and the performance when using the original tweet. The second ablation experiment verifies the performance of LLM when using different numbers of tweets. In the second ablation experiment, not all users can collect more than three tweets, so we only use users with more than or equal to five tweets collected in the stance detection dataset, and then randomly select one to five tweets from these users to input into LLM to evaluate the performance. In the one-tweet experiment, the input tweet can be different from the tweets in the original dataset. 

\subsection{Evaluation Metrics}
Similar to previous work, we calculate the label error rate $R_e$ according to (1). 
\begin{equation}
R_e = S_e/S_t
\end{equation}
$S_e$ is the number of error samples, and $S_t$ is the total number of samples. Previous work \cite{mohammad2016semeval,nakov2019semeval} uses the average F1 value of positive and negative samples to evaluate model performance. However, we find that in some targets, the number of positive or negative samples that can still be accessed is very small, and directly using the average F1 value will cause a large bias. Thus we use the $Accuracy$ for evaluation. 
\begin{equation}
Accuracy = S_c/S_t
\end{equation}

$S_c$ is the number of samples predicted correctly. Since a user may have multiple tweets in the original dataset, each one may be annotated with a different label, we calculate $R_e$ and $Accuracy$ based on the number of tweets rather than the number of users. This allows us to better compare our method with the original dataset.

\begin{table}[t]
\centering
\small
\setlength{\tabcolsep}{7pt}
\renewcommand{\arraystretch}{1.08}
\caption{Error rate statistics across different datasets.}
\label{tab:error_rate}
\begin{tabular}{llrrr}
\toprule
Dataset & Target & \# Posts & \# Errors & Error Rate \\
\midrule
\multirow{5}{*}{Stance}
& AT  & 133 & 30  & 29.3\% \\
& CC  & 114 & 22  & 19.3\% \\
& FM  & 114 & 50  & 43.9\% \\
& LA  & 130 & 54  & 41.5\% \\
& All & 491 & 156 & 31.7\% \\
\midrule
Sentiment  & All Topics & 120 & 28 & 23.3\% \\
HateXplain & All Topics & 76  & 37 & 48.7\% \\
\bottomrule
\end{tabular}
\end{table}

\section{Assessing Label Errors}

According to our guidelines, we evaluated the labeling errors of the three datasets. In the stance detection dataset, there were 156 tweets with incorrect labels. The error rate was as high as 31.7\%. Among them, the error rates of Atheism, Feminism, and Legalization of Abortion were as high as 29.3\%, 43.9\% and 41.5\% respectively. In the topic-based sentiment analysis dataset, the error rate is 23.3\%. In the hate speech detection dataset, the error rate is 48.7\%. This may be related to the unclear standards when the dataset was constructed and the fuzzy boundaries between the three categories of labels. Table \ref{tab:error_rate} shows the tweets and error rates in different targets or topics in the three datasets. 

We then perform a qualitative analysis of the errors in these labels. In the stance detection dataset, the target of Legalization of Abortion, a hashtag $\#repealthe8th$ repeatedly appears, which often means that the user is Irish and opposes the Eighth Amendment to the Irish Constitution. The Eighth Amendment to the Irish Constitution is a law against the Legalization of Abortion. Opposing the law means that the user's stance on the Legalization of Abortion is \textsc{Favor}. However, in the original dataset, a large number of tweets are annotated as \textsc{Against}. This is most likely because the annotators are not Irish and do not understand Irish culture and politics. In hate speech detection, mistakes often arise when annotations ignore individual-level context and pragmatic factors (e.g., speaker intent, in-group vs. out-group usage, quotation/irony, and the conversational setting). For instance, posts containing racial slurs may be labeled differently depending on whether the term is used to target a protected group or appears in reclaimed, self-referential, or quoted contexts; failing to account for such context can blur the boundary between “hate speech” and “offensive language,” illustrating the complexity of reliable annotation. More examples are given in Figure \ref{error_simples}. 

\section{Evaluation on Expanded Datasets}

\subsection{Quantitative Analysis}
We first evaluate the performance of the three models on the new datasets. As shown in Table \ref{tab:label_correction_results}, GPT-4o and Llama-3 show good performance on stance detection and sentiment analysis tasks. PHI-4 performs slightly worse on the stance detection task; However, if we use the original labels (uncorrected dataset labels, OL) for evaluation, the accuracy of the three models will drop significantly. The three models perform slightly worse in the hate speech detection task, but are still better than the accuracy using the original labels. This shows that label errors in the original dataset will seriously affect the evaluation of model performance. 

\begin{table}[t]
\centering
\footnotesize
\setlength{\tabcolsep}{4.2pt}
\renewcommand{\arraystretch}{1.08}
\caption{Model performance after user-context-aware relabeling. OL denotes evaluation using the original labels. SD: stance detection; SA: sentiment analysis; HSD: hate speech detection.}
\label{tab:label_correction_results}
\begin{tabular}{lrrrrrr}
\toprule
\multirow{2}{*}{Model} 
& \multicolumn{2}{c}{SD} 
& \multicolumn{2}{c}{SA} 
& \multicolumn{2}{c}{HSD} \\
\cmidrule(lr){2-3} \cmidrule(lr){4-5} \cmidrule(lr){6-7}
& Relabeled & OL & Relabeled & OL & Relabeled & OL \\
\midrule
GPT-4o     & 92.46 & 64.77 & 89.17 & 78.33 & 65.79 & 46.05 \\
Llama3-70B & 87.58 & 64.56 & 92.50 & 78.33 & 60.53 & 38.16 \\
PHI-4      & 79.02 & 61.51 & 92.50 & 76.67 & 53.95 & 35.52 \\
\bottomrule
\end{tabular}
\end{table}

\subsection{Ablation Studies}

We conduct two ablation experiments to evaluate the validity of multiple tweets from the same user. Table \ref{tab:mt_st_comparison} shows the results of the first ablation experiment. When using only the original tweet, the accuracy of all LLMs drops. This proves the necessity of using individual factors. Different posts from the same individual can complement each other and enhance the accuracy of prediction.

\begin{table}[t]
\centering
\footnotesize
\setlength{\tabcolsep}{5.2pt}
\renewcommand{\arraystretch}{1.08}
\caption{Comparison between multiple-tweet and single-tweet input settings. MT denotes multiple tweets, and ST denotes a single tweet. SD: stance detection; SA: sentiment analysis; HSD: hate speech detection.}
\label{tab:mt_st_comparison}
\begin{tabular}{lrrrrrr}
\toprule
\multirow{2}{*}{Model} 
& \multicolumn{2}{c}{SD} 
& \multicolumn{2}{c}{SA} 
& \multicolumn{2}{c}{HSD} \\
\cmidrule(lr){2-3} \cmidrule(lr){4-5} \cmidrule(lr){6-7}
& MT & ST & MT & ST & MT & ST \\
\midrule
GPT-4o     & 92.48 & 69.25 & 89.17 & 74.17 & 65.79 & 63.16 \\
Llama3-70B & 87.58 & 74.13 & 92.50 & 72.50 & 60.53 & 59.21 \\
PHI-4      & 79.02 & 60.29 & 92.50 & 71.67 & 53.95 & 50.00 \\
\bottomrule
\end{tabular}
\end{table}

\begin{figure}[t]
    \centering 
    \includegraphics[width=1\textwidth]{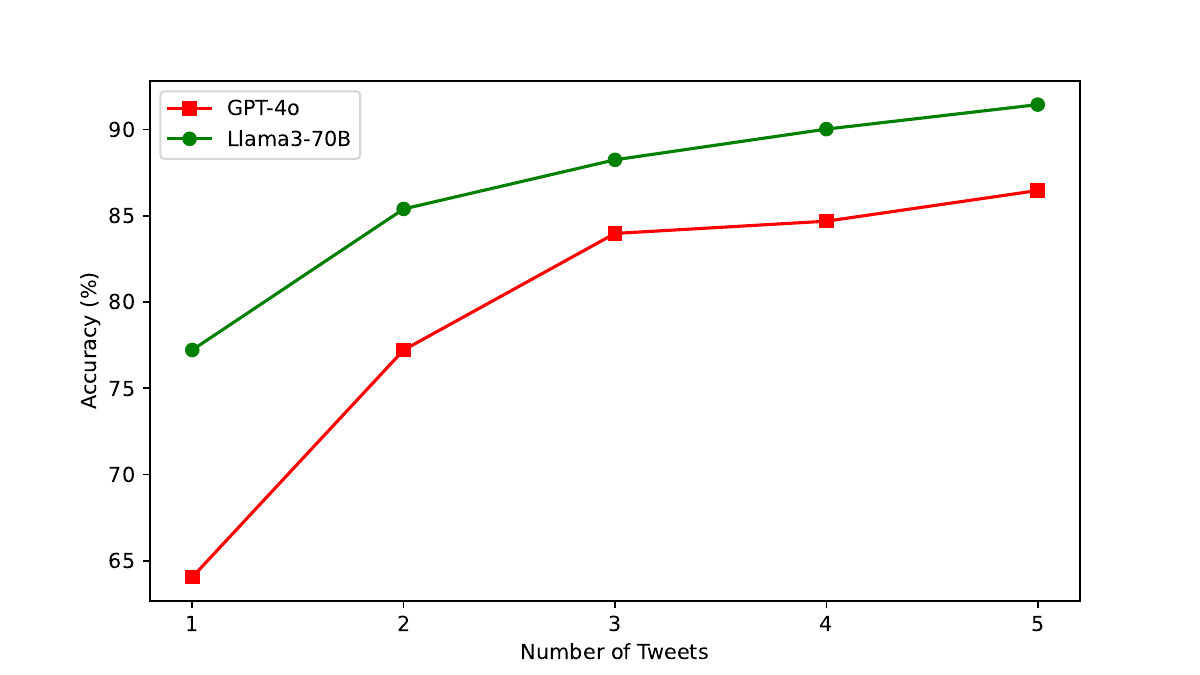} 
    \caption{Performance on the Semeval stance detection dataset using different numbers of tweets as LLM input.}
    ~\label{multi_tweets}
    
\end{figure}

Then we input LLM with one to five tweets from the same user in the stance detection task. We collected 281 users with more than five tweets, so we evaluated the effectiveness of multiple tweets on these 281 users. Since PHI-4 performed poorly before, we used LLama3-70B and GPT-4o for experiments. Figure \ref{multi_tweets} shows that three tweets can achieve good accuracy. Although the performance can continue to improve by increasing the number of tweets, the improvement is significantly reduced. Therefore, using three tweets is a choice that takes both performance and efficiency into consideration.


\subsection{Case Study}
\begin{figure}[tb]
    \centering 
    \includegraphics[width=1.1\textwidth]{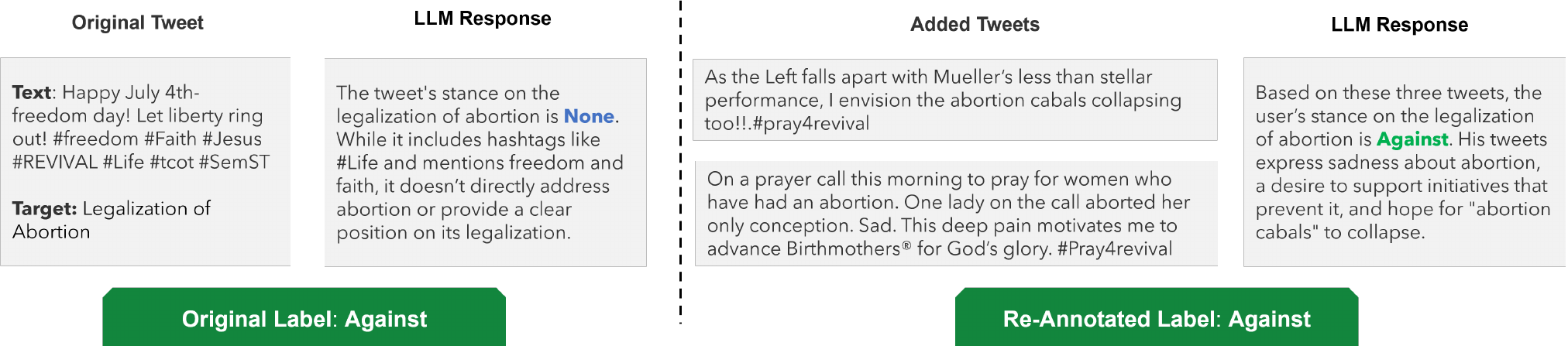} 
    \caption{Multi-posts example. In this case, although the user's stance is "Against" in both the new and original datasets, it is difficult or even impossible to infer the user's stance from the text in the original dataset. After adding other tweets from the user, LLM gives an accurate prediction.} ~\label{Case}
\end{figure}

We also conduct case studies of the results given by LLMs. We focus on two types of samples: The first type is samples where the new label is different from the original label. The second type is samples where the new label is the same as the original label, but the prediction results are different when using multiple tweets and a single tweet. We find that LLMs' explanations were basically consistent with the annotators' cognition.  A typical example is provided in Fig. \ref{Case}. The sample is annotated "Against" in both the new and original datasets, but even humans find it difficult to judge the user's stance on Legalization of Abortion through the original tweets. All three annotators also believe that the original tweet did not mention abortion at all, nor did it contain any clues supporting or opposing abortion. When only one tweet is used for stance detection, LLMs give a prediction result of "None", which is consistent with the annotator's cognition. After using the newly added two tweets, a total of three tweets for prediction, LLMs give the result of "Against". The three annotators also give the label "Against" based on the newly added tweets. This proves that expanding the dataset and increasing the information of the same user in the dataset is crucial for individual-level NLU. And further shows that annotation of individual tweets is prone to labeling errors. 

\section{Discussion and Conclusion}
Our research demonstrates the limitations of reducing individual-level NLU tasks to text-level tasks. The information lost in the reduction process not only leads to poor model performance but also causes annotators to misunderstand semantic information, resulting in a large number of label errors. Therefore, we call on dataset creators to fully consider social factors and reasonably choose guidelines to reduce systematic label errors when creating individual-level datasets in the future.

\section{Ethics Statement}
Our work on the datasets is conducted with a strong commitment to ethical principles. We prioritize privacy by collecting only publicly available tweets and strictly adhering to relevant guidelines for annotation and dataset sharing. In our research, we comply with the X Developer Agreement and Policy, ensuring that all content is used solely for academic research purposes. Tweets can only identify online users, not real individuals. Furthermore, we respect diverse religious beliefs and political perspectives.

Additionally, our research does not diminish the contributions of previous dataset creators; rather, we deeply appreciate their efforts. The datasets they developed serve as the foundation of our work.

\section{Acknowledgment}

This material is based upon work supported by NSF awards (IIS-2506643, and POSE-2346158). The views and conclusions contained in this document are those of the authors and should not be interpreted as necessarily representing the official policies, either expressed or implied, of the National Science Foundation and the US National Institutes of Health.

\clearpage
\appendix

\section{Keywords in searching}
In topic-based sentiment analysis and hate speech detection, we directly use topics as the keyword for search. In stance detection, the keywords are:
\begin{itemize}
    \item Atheism: $Atheism$, $God$, $Pray$
    \item Climate Change is a Real Concern: $Climate$, $Globalwarming$
    \item Feminist Movement: $Women$, $Feminism$, $Feminist$
    \item Legalization of Abortion: $Abortion$, $Women$, $Legal$
\end{itemize}

\section{Prompts Design}
We use a very simple prompt:

\begin{itemize}
    \item For stance detection:

    Read the question, provide your answer, and your confidence in this answer. Please make sure that the confidence level of your answers adds up to 1. Only output confidence levels. Do not output any other things. Please decide the following users' stance on the $Target$: Is it FAVOR, AGAINST, or NONE? These tweets are from the same user, so please assume they have the same stance. $Tweet1$ $Tweet2$ $Tweet3$  
    
    \item For topic-based sentiment analysis:

    Read the question, provide your answer, and your confidence in this answer. Please make sure that the confidence level of your answers adds up to 1. Only output confidence levels. Do not output any other things. Please decide the following users' sentiment on the $Topic$: Is it POSITIVE or NEGATIVE? These tweets are from the same user, so please assume they have the same sentiment. $Tweet1$ $Tweet2$ $Tweet3$

    \item For hate speech detection:

    The definition of hate speech and offensive speech is as follows: hate speech refers to offensive discourse targeting a group or an individual based on inherent characteristics (such as race, religion or gender) and that may threaten social peace. Note that hate speech can only be directed at individuals or groups of individuals. It does not include communication about States and their offices, symbols or public officials, nor about religious leaders or tenets of faith. Please consider both the individual factors of the publisher and social consensus before judging it as hate speech. For example, if a black person uses 'Nigger', it is usually not judged as hate speech, but if a white person uses 'Nigger', it is considered hate speech. 
    
    Offensive speech is defined as speech that makes others feel uncomfortable, insulted, or degraded, but does not necessarily have the consensus intention of systematic discrimination or inciting hatred. When judging offensive speech, it is necessary to consider social consensus, that is, whether the tweet is considered appropriate by the majority of people or a certain group in society. For example, some black people do not agree with the use of the word 'Nigger' among black people, then even if the publisher is black, please Judge the speech as offensive speech. 
             
    Other speeches are normal speech.

    Please decide whether this tweet is hate speech, offensive speech, or normal speech. Only output hatespeech, offensive or normal, don't output any other things: $Tweet1$.
    
    You can use these two tweets to help you make your judgment, they are other tweets posted by the same user: $Tweet2$ $Tweet3$
     
\end{itemize}

\section{Average F1 value}
In previous work, the average F1 value of positive and negative samples was often used to evaluate model performance of stance detection and sentiment analysis. The formula is as follows:
\begin{equation}
F_{avg} = \frac{F_P+F_N}{2}
\end{equation}
$F_P$ is the F1 value of the positive sample, and $F_N$ is the F1 value of the negative sample. In stance detection, positive samples are samples with the label $Favor$, and negative samples are samples with the label $Against$.

\begin{table}[tb]
    \centering
    \resizebox{\linewidth}{!}{
    \begin{tabular}{c|cc|cc}
        \hline
        ~ & \multicolumn{2}{c|}{Stance Detection} & \multicolumn{2}{c}{Sentiment Analysis}\\
        \hline
        Model & Total & Total (OL) & Total & Total (OL) \\
        \hline
        GPT-4o & 94.36 & 68.61 & 88.44 & 77.08 \\
        LLama3-70B & 90.29 & 68.94 & 92.06 & 76.67 \\
        PHI-4 & 78.87 & 65.96 & 92.06 & 74.88 \\
        \hline
    \end{tabular}
    }
\caption{The average F1 value of different models on the dataset after label correction. OL means original labels, which is the label of the original datasets without correction.} 
\end{table}

\clearpage

\clearpage
\section{More Error Samples and LLM Responses}

\begin{figure}[H]
    \centering
    \includegraphics[
        width=\textwidth,
        height=0.8\textheight,
        keepaspectratio
    ]{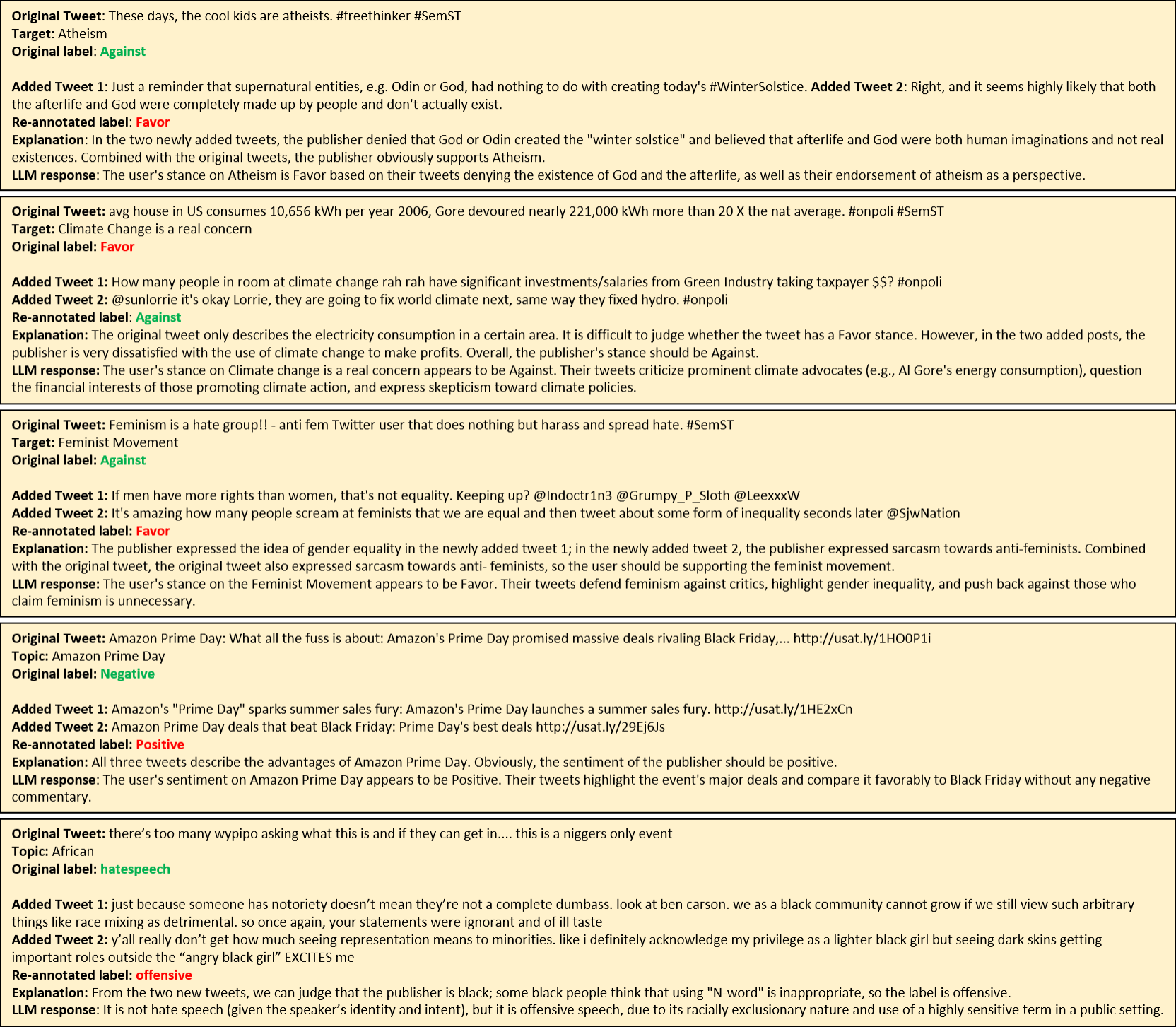}
\end{figure}

\bibliographystyle{splncs04}
\bibliography{custom}

\end{document}